\documentclass[hidelinks,11pt,letterpaper]{article}
\pdfoutput=1
\usepackage{emnlp2016}
\usepackage{times}
\usepackage{latexsym}
\usepackage{graphicx}
\usepackage{tikz}
\usetikzlibrary{arrows}
\usepackage{caption}
\usepackage{subcaption}
\usepackage{amsmath}
\usepackage{algorithm}
\usepackage[noend]{algpseudocode}
\usepackage{color,soul}
\emnlpfinalcopy

\def\emnlppaperid{***}

\title{Implementing a Reverse Dictionary, based on word definitions, using a Node-Graph Architecture}

\author{Sushrut Thorat\\
	   Center for Mind/Brain Sciences\\
	    University of Trento\\
	    Rovereto, TN 38068, Italy\\
	    {\tt sushrut.thorat94@gmail.com}
	  \And
	Varad Choudhari\\
  	Department of Computer Science\\
  	Rajarambapu Institute of Technology\\
  	Islampur, MH 415414, India\\
  {\tt varad.choudhari@gmail.com}}

\date{\today}

\begin{document}

\maketitle

\begin{abstract}
 
In this paper, we outline an approach to build graph-based reverse dictionaries using word definitions. A reverse dictionary takes a phrase as an input and outputs a list of words semantically similar to that phrase. It is a solution to the Tip-of-the-Tongue problem. We use a distance-based similarity measure, computed on a graph, to assess the similarity between a word and the input phrase. We compare the performance of our approach with the Onelook Reverse Dictionary and a distributional semantics method based on \emph{word2vec}, and show that our approach is much better than the distributional semantics method, and as good as Onelook, on a $3$k lexicon. This simple approach sets a new performance baseline for reverse dictionaries.\footnote{\scriptsize {In the proceedings of the 26th International Conference on Computational Linguistics (COLING 2016), pages 2797-2806. The test data and a demo code can be found at: https://github.com/novelmartis/RD16demo}} \nocite{bilac:04}
 
\end{abstract}

\section{Introduction}

A \emph{forward dictionary} (FD) maps words to their definitions. A \emph{reverse dictionary} (RD)~\cite{Sierra:00}, also known as an \emph{inverse dictionary}, or \emph{search-by-concept dictionary}~\cite{Calvo:16}, maps phrases to single words that approximate the meaning of those phrases. In the Oxford Learner's Dictionary\footnote{\scriptsize {Accessed: February, 2016}}, one definition of `brother' is `a boy or man who has the same mother and father as another person'. A reverse dictionary will map not only this phrase to `brother', but also phrases such as `son of my parents'. A reverse dictionary is primarily a solution to the \emph{Tip of the Tongue} \nocite{Schwartz:99}problem~\cite{Schwartz:11} which regularly plagues people when they want to articulate their thoughts. It can also be used in the treatment of \emph{word selection anomic aphasia}~\cite{Rohrer:08}, a neurological disorder in which patients can identify objects and understand semantic properties, but cannot name the object or produce one word to describe the concept.

Popular languages let us create a multitude of phrases from a finite number of words. A static database of all possible phrases is unbound, if not infinite~\cite{Ziff74}. We need to dynamically compute the output word(s) from the input phrase. To map a phrase to a word, we have to compute the meanings of the phrase and the word~\cite{Fromkin:11}.  The \emph{principle of compositionality} states that the meaning of an expression is composed of the meaning of its parts and the way they are combined structurally. The most basic parts, words, can be defined in terms of word definitions, references to objects, or lexical relations and hierarchies. Computing the meaning of a phrase requires constructing the constituent tree and recognising the relationship between the constituents, which is a complex, open problem.

Compositional Distributional Semantic Models have been used towards computing the meaning of a phrase, with partial success~\cite{Baroni:13,Erk:12}. Recurrent neural networks show promise in learning continuous phrase representations. They are used towards syntactic parsing beyond discrete categories such as NP and VP, in an attempt to capture phrasal semantics~\cite{socher:10}. A recent work has used neural language embedding models (RNNs with LSTMs) to understand phrases by embedding dictionaries~\cite{hill:15}. But it doesn't perform exceptionally better than the existing reverse dictionaries (OneLook, etc.)

If we are to ignore the ordering of words in a phrase, the performance of such a system would not be maximal. But we could then work just with well-studied lexical relations. Research into building reverse dictionaries has mainly focused on lexical relations than the structural or contextual combination of words. The attempts in defining a \emph{similarity measure} between words have been summarised in~\cite{Rada:06}. An attempt towards situational understanding and contextual selection of words can be seen in~\cite{Granger:82}. The creation of WordNet~\cite{Miller:95} boosted the use of lexical relations and hierarchies, as in ~\cite{Dutoit:02,Kahlout:04,Shaw:13,Mendez:13,Calvo:16}. Most of these approaches extract \emph{input words} from the input phrase and expand their search through lexical relations and hierarchies, towards a similarity measure between the phrase and the words. \cite{Zock:08} take inspiration from human word synthesis and implement a user-guided search to the desired word. All these approaches have achieved partial success, but the problem stays unsolved.

We explore the possibility of using word definitions towards establishing semantic similarity between words and phrases. Definitions are dense sources of semantic information about words (which makes it difficult to extract information from them without using exact syntactic structures such as constituent trees), and in our approach, we employ them exclusively. We assume that the significance of the meaning of a word to a definition is proportional to its frequency throughout the definitions in the FD. We extract the meaning from the \emph{content words}~\cite{Fromkin:11} contained in the phrase.  We split the input phrase into these component \emph{input words}, implement a graph-search through related words (relation through definition), and use a \emph{distance-based} similarity measure to compute words which represent the meaning of the input phrase. A graph encodes the word relations in its connectivity matrix, on which the similarity measures are computed. We detail our approach next.

\section{System Description}

\begin{figure}[!h]
\centering
\includegraphics[width=0.47\textwidth]{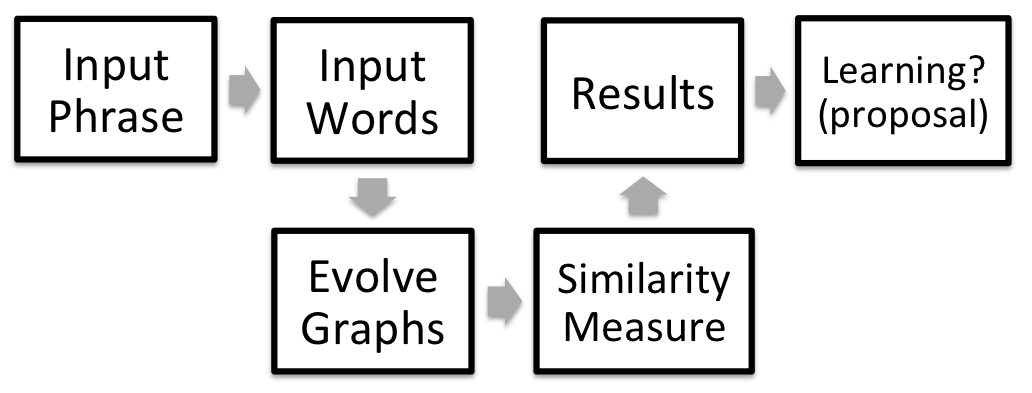}
\caption{Operation of the Reverse Dictionary. The graphs' connectivities are based on the \emph{reverse map}, a concept we will introduce shortly.}
\label{fig:operate}
\end{figure}

The block diagram of the operation of the RD is depicted in Fig.~\ref{fig:operate}. We now discuss the concept of the reverse map, central to the structure of our graph, and the process of obtaining the connectivity matrix underlying our graph.

\subsection{The Reverse Map}

\begin{figure}[!h]
\centering
\includegraphics[width=0.4\textwidth]{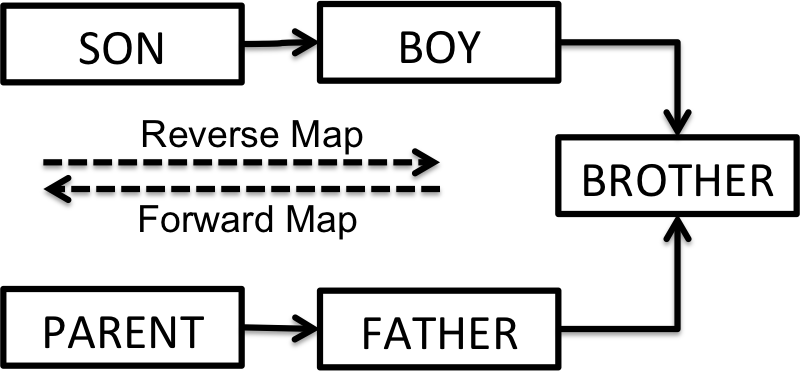}
\caption{Each solid arrow indicates a \emph{in the definition of} relation. This is the reverse map leading the phrase `Son of my parents' to the word `brother', after extraction of the input words. Note that this is one of the many sets of connections to all words on the graph from that phrase.}
\label{fig:maps}
\end{figure}

In a \emph{forward map}, words branch out to the words that are contained in their definitions. In a \emph{reverse map}, words branch out to the words whose definitions they are contained in. An example of a reverse map\footnote{\scriptsize {Based on the definitions from the Oxford Learner's Dictionary.}} is shown in Fig.~\ref{fig:maps}.

If the input phrase is a definition, a search depth of one (branching out from the words of the input phrase to the definitions they are contained in) of the reverse map will lead to the required word. A search depth beyond one provides us with semantic information about words whose definitions encompass or relate to concepts that encompass or relate to the input words, and so on. Increasing search depth obscures the relationship between words, which is the basis for the definition of our similarity measure. A reverse map suggests semantic convergence in a shallow search, although the convergence might occur on multiple words, which is acceptable as they might be semantically-similar. Intuitively, a forward search seems to `fragment' the meaning of the input word, and is expected to perform worse than the reverse search in defining word relationships in our approach.

\subsection{Connectivity Matrix of the Reverse Map}

The steps in the construction of the connectivity matrix, based on the reverse map, are as follows. Our inputs are a forward dictionary, a list of functional words, and a lemmatizer. We process the forward dictionary to retain content words in their base form. We then construct the forward-linked list, transform it into a back-linked list, and then construct the back-linked connectivity matrix. Similarly, we can also construct the forward-linked connectivity matrix.

\subsubsection{Processing the Forward Dictionary}

A forward dictionary (FD) can be viewed as a two-dimensional list. The rows in the first column contain the words, and the rows in the second column contain the corresponding definitions. We reduce all words in column one to their lemmas, their base forms\footnote{\scriptsize Using the \emph{pattern lemmatizer}~\cite{Smedt:12} and \emph{wordnet morphy}~\cite{Bird:09}.}. We then delete all the \emph{functional words}\footnote{\scriptsize {The functional words were obtained from Higgins, $2014$: http://myweb.tiscali.co.uk/wordscape/museum/funcword.html}}~\cite{Fromkin:11}, and the corresponding definitions in column two. For our purposes, we pool all the definitions of a particular word into a single cell, parse them through the lemmatizers and delete all the functional words within them. We term the resulting list the \emph{forward-linked list}. We now generate the \emph{back-linked list}.

\subsubsection{The Back-linked list}

We number the words in column one of the forward-linked list in the alphabetical order (word-id). We substitute all the words in column two by their word-ids. The back-linked list is generated by the following algorithm:
\begin{align*}
    &\textbf{for } i \textbf{ in } [1,\textbf{length}(\text{Fs})] \textbf{ do}:\\
    &\quad\textbf{for } j\textbf{ in } \text{Fs}(i,2) \textbf{ do}: \text{Bs}(j,2).\textbf{append}(i)
\end{align*}
where \emph{Fs} is the forward-linked list, and \emph{Bs} is the back-linked list. We created a list which points a word to the words whose definitions it lies in.

\subsubsection{The Back-Linked Matrix}
\label{sec:blm}

We now generate the matrix which represents the connections (weights) between the nodes (words, in this case). The \emph{back-linked matrix} (BLM) is generated by the following algorithm:
\begin{align*}
&\textbf{for } i\textbf{ in } [1,\textbf{length}(\text{Fs})]\textbf{ do}:\\
&\quad\textbf{for } j\textbf{ in } \text{Bs}(i,2)\textbf{ do}:\text{BLM}(j,i)=1\\
&\quad\text{BLM}(i,i) = 0
\end{align*}

We will see in section~\ref{sec:gra} that many words in a dictionary do not appear in any definition, and so cannot contact all words in the wordlist through the reverse map. But we would like to obtain the similarity measure between any two words in the wordlist. As a simple measure in ensuring complete connectivity, we build a \emph{mixed back-linked matrix} (mBLM) which has forward-linked connections for words that do not have sufficient back-linked connections. The mBLM is generated by the following algorithm:
\begin{align*}
&\text{mBLM}=\text{BLM}; l =\textbf{length}(\text{BLM})\\
&\textbf{for } i\textbf{ in } [1,l]\textbf{ do}:\\
&\quad \text{S}=\textbf{zeros}(l,1); \text{S}(i,1)=1; \text{S}=\text{BLM}^{p}\otimes \text{S} \\
&\quad\textbf{if } \text{S} \neq \textbf{ones}(l,1)\textbf{ then do}: \\
&\qquad\text{mBLM}(:,i) = \text{mBLM}(:,i)+\text{BLM}(i,:)^\textbf{T}\\
&\qquad\textbf{for }j\textbf{ in } \text{mBLM}(:,i)\textbf{ do}: \\
&\quad\qquad\textbf{if }j > 0\textbf{ then do}: j = 1 \\
&\quad \text{mBLM}(i,i) = 0
\end{align*}
where\footnote{\scriptsize {\emph{A(:,i)} denotes the \emph{i}\textsuperscript{th} column of \emph{A}. \emph{A}\textsuperscript{\textbf{T}} denotes the transpose of \emph{A}.}} $S$ is a dummy variable (denoting the states of the nodes - introduced as such in section~\ref{sec:nga}), and \emph{p} is a parameter which corresponds to the maximal depth of search required to compute the connectivity of the graph (which corresponds to the maximum non-redundant search through the graph, which is mentioned in section~\ref{sec:gra}). We will assess in section~\ref{sec:perf} the change in performance by the inclusion of the said forward links.

\subsection{The Node-Graph Architecture}
\label{sec:nga}

The connections between the nodes in our graph are given by the BLM. Each word is represented by a node. Each node has two states $\left \{  0,1\right \}$. They respond to incoming signals, by processing their state and passing the signal to downstream nodes. If $S$ is the state of the population of nodes, $n$ denotes the number of time steps to be taken, $I_{in}$ denotes the input signals to the nodes, $I_{ext}$ denotes the external bias signals to the nodes, and $I_{out}$ denotes the output signals from the nodes, then the dynamics of the states of the nodes are computed by the algorithm:
\begin{align*}
&t=0\\
&\textbf{while } t \leq  \text{n}\textbf{ do}:\\
&\quad \text{I}_{out} = \text{S}\\
&\quad \text{I}_{in} = \text{BLM}\otimes \text{I}_{out} + \text{I}_{ext} \\
&\quad \textbf{for } i\textbf{ in } [1,\textbf{length}(\text{BLM})]\textbf{ do}: \\
&\qquad \textbf{if }\text{I}_{in}(i)\geq 1\textbf{ then do}: \text{S}(i) = 1\\
&\qquad \textbf{if }\text{I}_{in}(i)== 0\textbf{ then do}: \text{S}(i)=0\\
&\quad t = t+1
\end{align*}

We create a graph for each \emph{input word} (we obtain these from the input phrase by parsing it through the same operations as the definitions), and turn the input currents to their corresponding word-ids in their corresponding sheets to $1$ (at $t=0$ using the I$_{ext}$ bias term). Then we let the graphs evolve with increasing $t$. We are, in effect, expanding the tree of words to be able to effectively implement the similarity measure. $n$ also represents the depth of the search. We term the evolution of $S$ until the step $n$ a \emph{n-layered search}.

\subsection{The Similarity Measure}
\label{sec:sim}

We use a distance-based measure of similarity.

We define the distance $d_{Y,X}$ from a word $X$ to another word $Y$ as the depth of search required to evolve a state with only $S_X = 1$, to the first state with $S_Y = 1$. 
Note that $d_{Y,X}\neq d_{X,Y}$.

We calculate the frequencies of appearances, $\left \{ \nu_Z \right \}$ throughout definitions, for all words $\left \{ Z \right \}$ in the wordlist.

We define the similarity measure $E_{W,P}$ of a word $W$ to an input phrase $P$ containing
the input words $\left \{ P_i \right \}$ as: 
\begin{equation*}
\text{E}_{W,P} =\frac{\sum _i \left ( \nu_{P_i}\times d_{W,P_i} \right )^{-1}}{\sum _i \nu_{P_i}^{-1}}
\end{equation*}

We weighted the inverse of the distances between the words with the inverse of the frequencies of the input words. So, the similarity measure includes a measure of `semantic importance' of each input word in the input phrase. We calculate the similarity measure of each word to the input phrase, and output the words in the decreasing order of similarity. As every word is connected to every other word in the reverse map given apt search depth, the similarity measure becomes important in finding relevant output. Our similarity measure states, the smaller the distances from the input words, the more similar is the word to the input phrase. Minimal distances ensure that the semantic similarity remains meaningful.

\subsection{System Summary}
\label{syssum}

The user inputs a phrase. Input (content) words are extracted from the phrase. Graphs are generated for each input word, and in each graph, the node corresponding to the input word is activated. The graphs are evolved to the \emph{maximum non-redundant} search depth (see section~\ref{sec:gra}). The similarity measure, to the input phrase, is computed for every word in the lexicon, and the words are ranked according to their similarity measures, leading to the output.

\section{Graph exploration}
\label{sec:gra}

We construct BLMs and mBLMs based on the processed\footnote{\scriptsize {Words which appeared in Oxford Learner's dictionary definitions, but were not part of the wordlist, were added to the wordlist for consistency. The modified wordlist contains $3107$ words, and is referred to as $3$k, in this paper.}} Oxford $3000$ wordlist\footnote{\scriptsize {http://www.oxfordlearnersdictionaries.com/about/oxford3000}}, and a BLM for the entire WordNet~\cite{Miller:95} lexicon (WL). We use the Oxford Learner's dictionary (OLD), Merriam-Webster dictionary\footnote{\scriptsize {Accessed: February, 2016}} (MW), and WordNet (WN) as forward dictionaries for the Oxford $3000$ wordlist, and WordNet for the WordNet lexicon (WL). We also build a BLM and a mBLM by pooling definitions (Fusion BLM) from the three forward dictionaries, for the $3$k wordlist, to check the effect of using multiple dictionaries on performance.

Before we move on to analyse the performances, let's look deeper into the connectivity matrices we generated. All the BLMs and mBLMs are sparse\footnote{\scriptsize {Sparsity (proportion of $0$'s in the matrices): $0.99$ ($3$k Fusion BLM), and $0.99$ (WordNet lexicon BLM)}}. We use the \emph{compressed sparse row} format from SciPy~\cite{Jones:01} to store and process our matrices. 

\begin{figure}[!h]
\centering
\includegraphics[width=0.5\textwidth]{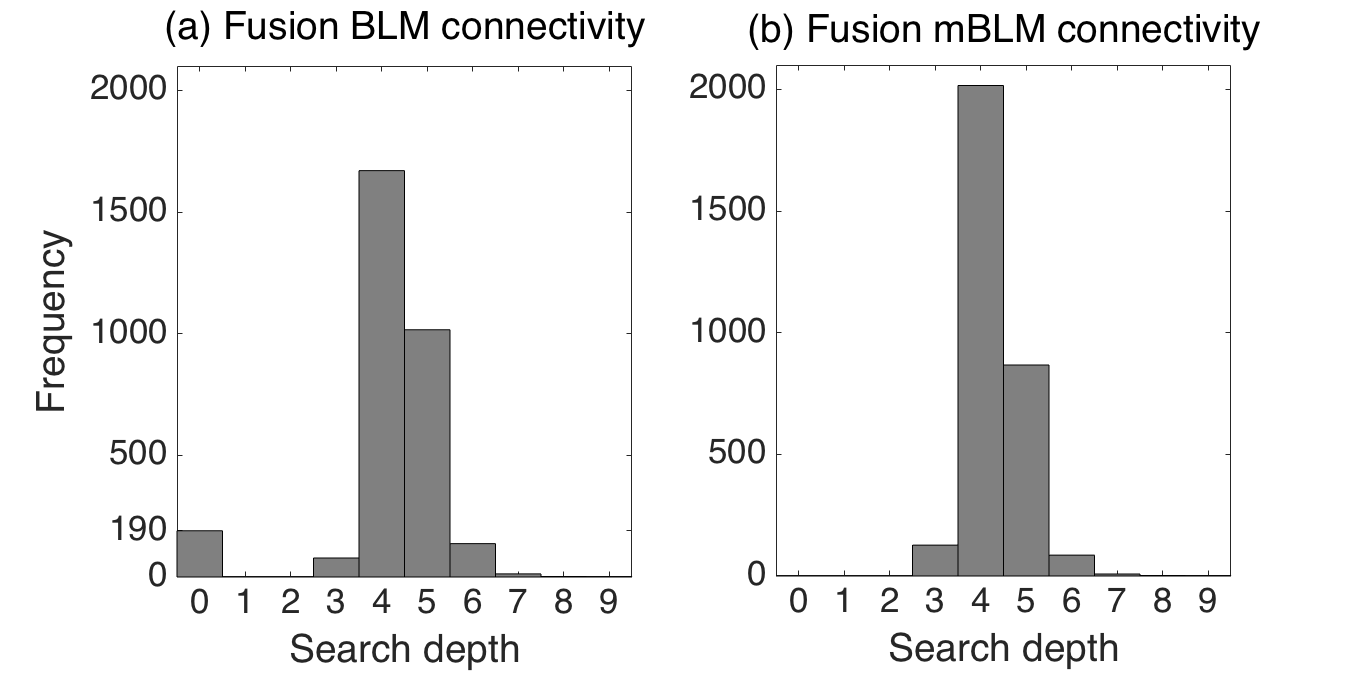}
\caption{Distribution of the minimum search depth required by a word to excite the entire graph. If a word does not excite the entire graph, a value of zero is assigned to its minimum search depth.}
\label{fig:fusion}
\end{figure}

In the $3$k wordlist case, the number of connections in the Fusion BLM is greater than the BLMs built with individual FDs. In Fig.~\ref{fig:fusion}(a), we see that there are $190$ words which cannot connect to the entire graph through the Fusion BLM. So, we build a mBLM, as proposed in section~\ref{sec:blm}, and ensure complete connectivity of the graph, as seen in Fig.~\ref{fig:fusion}(b). As all words can connect to all other words in $9$ steps at the most, a search depth greater than $9$ would be redundant when we use the Fusion BLM. The \emph{maximum non-redundant} search depths for the individual BLMs are as follows: $11$ (OLD), $9$ (WN), and $11$ (MW).

\begin{figure}[!h]
\centering
\includegraphics[width=0.5\textwidth]{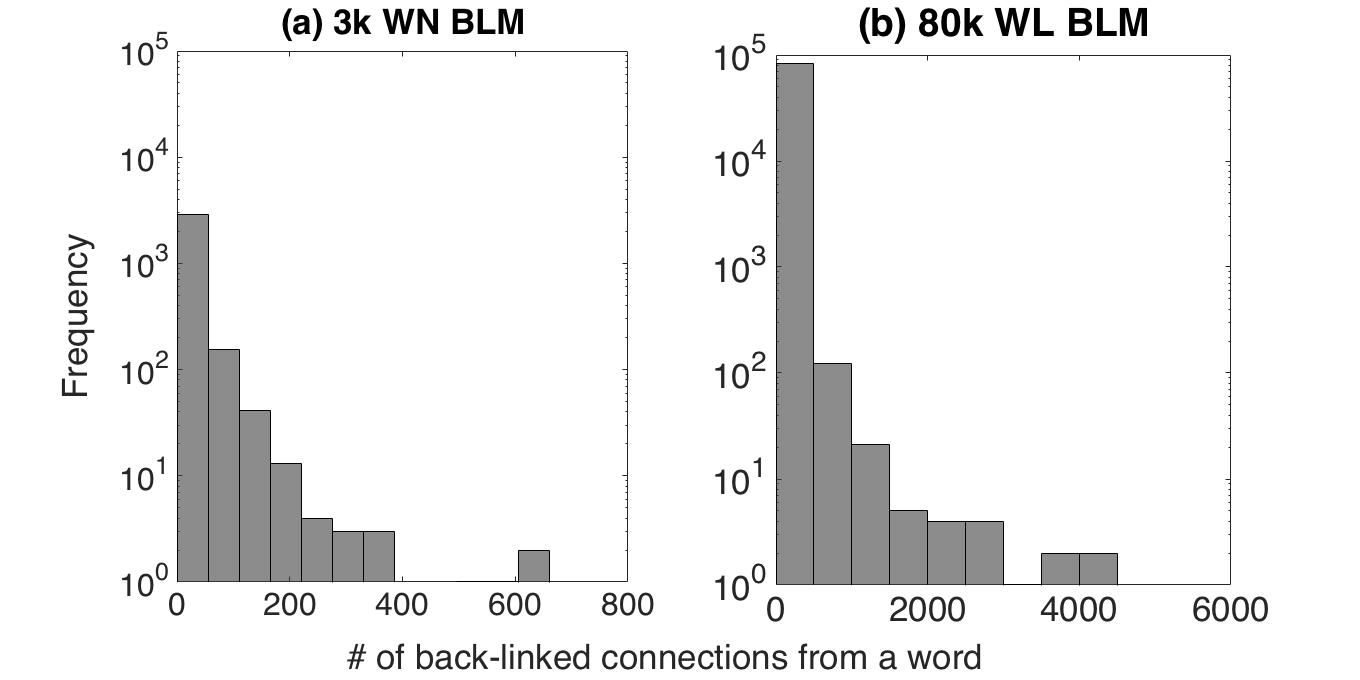}
\caption{Distribution of the number of back-linked connections from the words, in the reverse map.}
\label{fig:freq}
\end{figure}

The maximum required search depth for the WordNet lexicon BLM is $19$. $53,711$ words out of $82,603$ do not map to any word in the reverse map. Those words are infrequent in the language and are not used to define other words. Fig.~\ref{fig:freq}(b) depicts the distribution of the number of back-linked connections from the words in the reverse map for the $80$k WL BLM ($\mu=7.81$, $\sigma=62.86$, max $=6163$), as compared to the distribution for the $3$k WN BLM ($\mu=18.10$, $\sigma=36.14$, max $=615$) in Fig.~\ref{fig:freq}(a). The huge number of backward-linked connections for some words in $80$k WL BLM would confound the accuracy of the similarity measures, and a drop in performance is expected.

\section{Performance Analysis}
\label{sec:perf}

The only available online reverse dictionary is the Onelook Reverse Dictionary~\cite{onelook}, with which we will compare our algorithm's performance. Onelook is a commercial application, and its architecture is proprietary. We know that it indexes $1061$ dictionaries and resources such as Wikipedia and WordNet. The lexicon of Onelook is much bigger than $3$k. In the performance comparison, we state the performance with (termed as `corr') and without adapting the outputs to the $3$k lexicon. 

We also compare our approach with a distributional semantics method, based on \emph{word2vec} which represents words as vectors in a linear structure that allows analogical reasoning. In that vector space, the vector `king + woman - man' will have a high similarity with the vector for `queen'~\cite{mikolov:13b,mikolov:13a}. We average the vector representations of input words, and search word vectors most similar to the resulting vector (cosine similarity). This is an established method of building phrase representations from word representations~\cite{mitchell:10}. The performance of such an approach\footnote{\scriptsize {Based on the implementation of $word2vec$ by Daniel Rodriguez at {https://github.com/danielfrg/word2vec}, trained on a corpus with $15.8\,$million words, and a vocabulary of $98$k.}} is shown in Table.~\ref{tab:performance} (as ‘W$2$V’).

\subsection{Performance Test}

\begin{table*}[!ht]
  \centering
  \begin{tabular}{| l || c | c | c || c | c | c |}
  \hline
  	Test Type $\rightarrow$ & \multicolumn{3}{|c|}{Macmillian Word Definitions (179)} & \multicolumn{3}{|c|}{User Concept Descriptions (179)} \\
  	\hline \hline
  	Evaluation $\rightarrow$ & \textbf{Accuracy} & Rank & Rank & \textbf{Accuracy} & Rank & Rank\\
  	Models $\downarrow$ & @$1/10/100$ & Median & $\sigma$ & @$1/10/100$ & Median & $\sigma$\\ \hline \hline
	Onelook  & $.19/.41/.65$ & $5$ & $24$ & $.04/.21/.40$ & $10$ & $26$\\ \hline
	\textbf{Onelook, \emph{corr}*}  & $.20/.46/.68$ & $3$ & $20$ & $.07/\mathbf{.26}/.52$ & $13$ & $30$\\ \hline  
	W$2$V  & $.01/.06/.20$ & $23$ & $30$ & $.01/.05/.18$ & $34$ & $28$\\ \hline
	W$2$V, \emph{corr}*  & $.02/.11/.29$ & $21$ & $26$ & $.01/.08/.26$ & $21$ & $29$\\ \hline
	Chance, $3$k  & $10^{-4}/10^{-3}/.03$  & $50$ & $29$ & $10^{-4}/10^{-3}/.03$ & $50$ & $29$\\ \hline \hline
	Fusion, FLM  & $.02/.10/.21$ & $12$ & $28$ & $.01/.07/.22$ & $16$ & $21$\\ \hline
	\textbf{Fusion, mBLM}  & $.25/\mathbf{.55}/\mathbf{.84}$ & $4$ & $22$ & $\mathbf{.10}/.23/\mathbf{.53}$ & $14$ & $26$\\ \hline
    OLD, mBLM  & $\mathbf{.26}/.52/.78$ & $4$ & $23$ & $.04/.17/.43$ & $14$ & $25$\\        \hline
    WN, BLM  & $.08/.27/.54$ & $11$ & $26$ & $.06/.18/.41$ & $14$ & $26$\\        \hline
    MW, mBLM  & $.17/.39/.63$ & $5$ & $20$ & $.05/.20/.43$ & $15$ & $25$ \\       \hline
    WL, $80$k  & $.03/.15/.36$ & $18$ & $26$ & $.05/.11/.24$ & $14$ & $25$\\ \hline
    WL, \emph{corr}*  & $.07/.26/.52$ & $10$ & $25$ & $.07/.18/.35$ & $10$ & $23$\\ 
    \hline
  \end{tabular}
  \caption{Performance of the various models. Accuracy @$n$ is the fraction of the phrases with the rank of the target word less than or equal to $n$, in their outputs. $\sigma$ is the standard deviation. Only the phrases with target words having ranks less than 100 were considered in calculating the median and variance. The $3$k cases (OLD, WN, MW, Fusion) were evaluated at a search depth of $11$, and the $80$k case (WL) at a search depth of $19$. *\emph{corr} indicates the cases where the outputs were truncated to fit in the $3$k lexicon, for fair comparison. (Note: Accuracy - higher is better; Rank median - lower is better; Rank variance - lower is better.)}
  \label{tab:performance}
\end{table*}

The reverse dictionary outputs multiple candidate words. We introduced users to the concept of the reverse dictionary and asked them to generate phrases they would use to get to a given word, if they would have forgotten the word but retained the concept. $25$ such users generated $179$ phrases, a sample of which is presented in Table.~\ref{tab:tester}. The performance is gauged by the ranks of the words in the outputs of their user-generated phrases\footnote{\scriptsize {An input phrase can have multiple semantically similar words. Analysing the semantic quality of each output would be the ideal test. This could be done using a function of the sum of the ranks of each output weighted with their distances (in a high-dimensional semantic space such as $word2vec$) from the target word. However, previous approaches have used just the rank of the target word (which is nevertheless a good indicator of performance), and here we do the same.}}. 

We also test all the approaches on one-line definitions for the $179$ words, obtained from the Macmillian Dictionary\footnote{\scriptsize {Accessed: May, 2016}}.

\subsection{Performance results}
\label{sec:perf_r}

\begin{figure}[!h]
\centering
\includegraphics[width=0.45\textwidth]{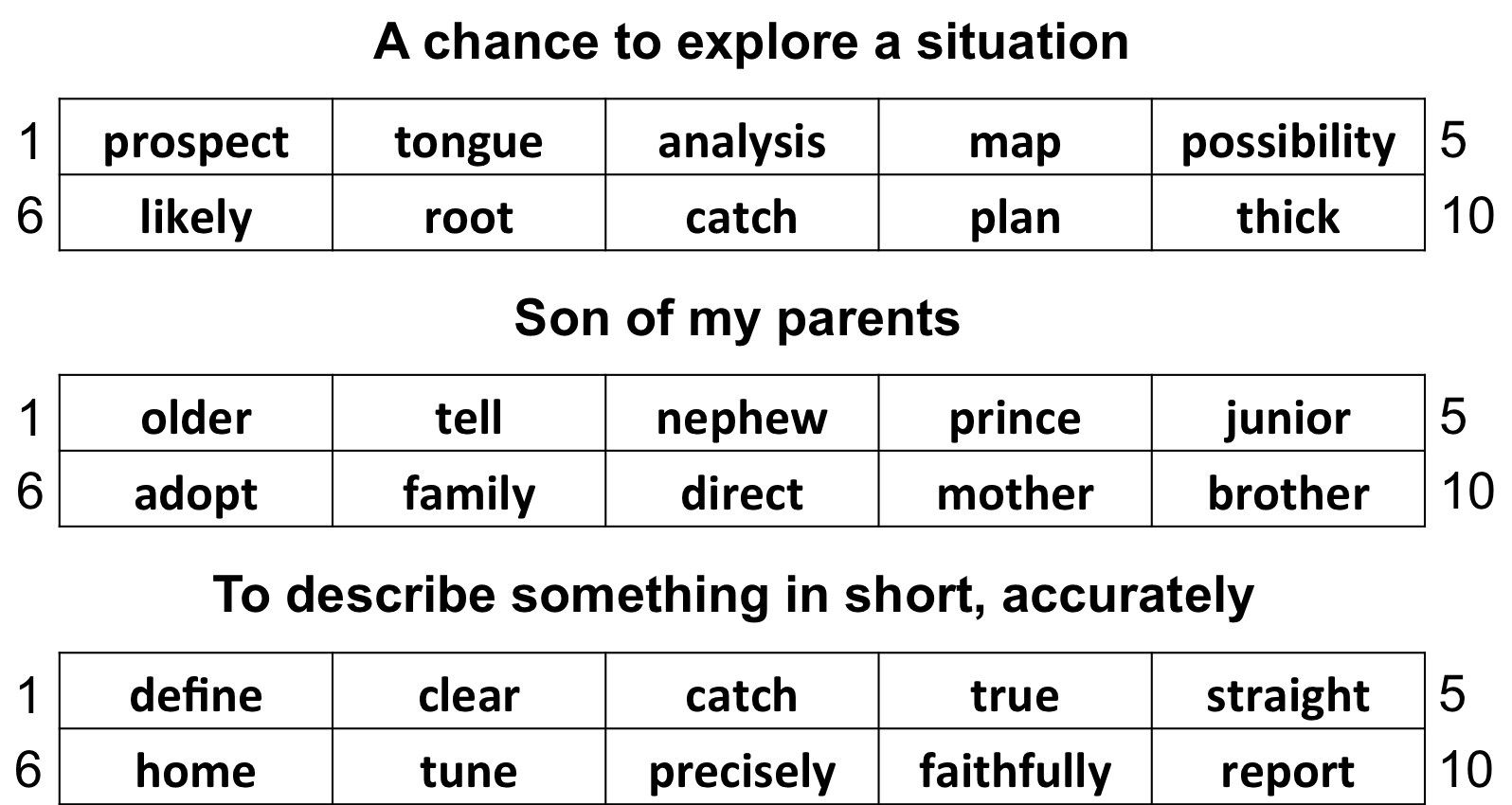}
\caption{The first $10$ words of the outputs obtained using the $3$k Fusion mBLM (n=$9$), for three input phrases.}
\label{fig:example}
\end{figure}

Example runs of the RD, using the $3$k Fusion mBLM, are presented in Fig.~\ref{fig:example}. The distributions of ranks, for the various BLMs/mBLMs (whichever has greater \% of ranks under $100$ for each case), \emph{word2vec}, and Onelook, are stated in Table.~\ref{tab:performance}. Onelook did not provide any outputs for $18$ phrases out of the $179$ user-generated phrases, and $72$ out of the $179$ definitions from the Macmillan dictionary. Instead of considering these as failures, we factor out these phrases while evaluating Onelook. The performance of all approaches is significantly better than chance, as seen through the comparison of performance with `Chance' which represents the expected values of performance for random rank assignments to the target words\footnote{\scriptsize The expected value of the accuracy @k, over random rank assignments, is given by: $\sum_{n=0}^{P_r} \frac{n}{P_r}.\frac{^{P_r}\textrm{C}_{n}k^n(N-k)^{P_r-n}}{N^{P_r}} = \frac{k}{N}$, where $P_r$ is the number of test phrases, and $N$ is the size of the lexicon.} (considering the $3$k lexicon). The cases of interest are highlighted in the table. 

\begin{table}[ht]
  \centering
  \begin{tabular}{| l || p{5cm} |}
  	\hline
  	Words & Phrases\\ \hline \hline
    Variation & A change or changes between two or more things \\        \hline
    Attractive & Something that is catchy\\  \hline
    Plus & The operation used to increase\\       \hline
    Church & Place to meet god\\ \hline
    Possession & Taking full control over a thing\\
    \hline
  \end{tabular}
  \caption{Sample user-generated phrases, used for testing the performance of the RD.}
  \label{tab:tester}
\end{table}

All the $3$k cases using a BLM/mBLM have a higher performance than the $3$k Fusion forward-linked matrix (FLM). Fusion of the individual $3$k BLMs yields better performance. The $3$k Fusion BLM performs at least as well as Onelook. The use of mBLMs is fruitful as they increase the performance in some cases.  The performance does not change much across search depth as seen in Table.~\ref{tab:perf_n}, suggesting that our approach works well even at a shallow search. Deeper search is required only when a phrase is semantically vague or non-specific, and markedly different from dictionary definitions. Both our approach and Onelook outperform the W$2$V approach. We conclude that our approach works well with a $3$k wordlist. Although the ranks' median and variance are indicative of the performance (hit rate, and robustness), they are marred by the accuracies, so we do not use them in our inferences. 

\begin{table}[!ht]
  \centering
  \begin{tabular}{| c || c | c | c | c |}
  	\hline
  	Accuracy $\downarrow$ & $n=1$ & $n=2$ & $n=3$ & $n=10$\\ \hline \hline
    @$1$ & $.08$ & $.07$ & $\mathbf{.10}$ & $\mathbf{.10}$\\        \hline
    @$10$ & $\mathbf{.25}$ & $.22$ & $.23$ & $.23$\\        \hline
    @$100$ & $\mathbf{.55}$ & $.52$ & $.53$ & $.53$\\ 
    \hline
  \end{tabular}
  \caption{Performance across search depth (n) for the user-generated phrases, in the Fusion mBLM case. The output becomes stable beyond a search depth of $3$. The search depth at which the output becomes stable varies with the BLMs.}
  \label{tab:perf_n}
\end{table}

However, the performance drops significantly when the entire processed WordNet lexicon (WL, $80$k) is the FD. The words that lie in the definitions of other words are a small subset of the WL wordlist. As seen in Fig.~\ref{fig:freq}, there are $163$ words in the WL wordlist which map to more than $500$ words in the reverse map. Therefore, the distances of multiple words to the input words are similar, obscuring the semantic content of the similarity measure. This is a potential limitation of our approach, for which there is no trivial fix. 

We also assessed the performance of the Fusion mBLM on the $200$ test phrases used in~\cite{hill:15}. The size of their lexicon is $66$k. We cannot upscale the outputs of our $3$k cases to $66$k, so a direct fair comparison with their results is not possible. However, we can downscale the outputs of Onelook (on the $200$ phrases) to $3$k and compare with it, thus providing an indirect comparison with the approach used by Hill et al. The @$1/10/100$ accuracies of the Fusion mBLM are $.16/.39/.62$. But $33$ target words do not lie in the $3$k lexicon. The accuracies excluding the corresponding phrases are $\mathbf{.19/.46/.74}$. The @$1/10/100$ accuracies of the Onelook (scaled to $3$k) are $.08/.21/.30$. But $101$ phrases do not return any outputs. The accuracies excluding those phrases are $\mathbf{.16/.42/.61}$. The accuracies of Onelook and the RNN approaches in Hill et al. are equivalent. We thus conclude that the performance of our approach is at least as good as the RNN approaches, on a $3$k lexicon.

\section{Recommendations}
\label{sec:disc}

The graph structure opens up a semantic dimension by letting us mutate the level of significance a word has in a definition, through the connectivity matrix. We can introduce this information in the similarity measure by scaling the weights of the connections between the words with distances equal to \emph{one}. The definitions provided in the dictionary cannot populate the new dimension. One could consider the use of semantic rules, or lexical relations, or user feedback. Such a learning algorithm could use further exploration.

There are multiple points in our approach which could use either improvement or emphasis. We use multiple graphs for calculating the similarity measure. This is done because we do not want the distance of a word from an input word to be a function of all the input words. Using \emph{Spiking Neural Networks}~\cite{ghosh:09}, we could implement the similarity measure using a single graph by frequency tagging the distances from each input word, although it isn't clear how much advantage it would confer in terms of performance. 

A matrix of pair-wise distances between all words could be used to evaluate the similarity measures, instead of evolving a graph. Such a matrix won't be sparse, and in the case of a $80$k lexicon would be $50\,$gigabytes in size (compared to $10\,$megabytes in CSR sparse format for the BLM), making it impractical to deploy the algorithm on mobile devices. Execution time and memory requirement are not a problem for our approach. Our approach is an easy and computationally cheap method of implementing semantic search with a graph, which performs at least as well as the Onelook reverse dictionary.

There is significant drop in performance when the WordNet $80$k lexicon is used (the mBLM doesn't help). The use of multiple forward dictionaries might boost the performance, as in the case of Fusion mBLM, but as mentioned in section~\label{sec:perf_r}, the branching factor of the graph is too high, obscuring the similarity measure. Although this might make the approach impractical, it does serve as a new baseline. A simple approach like ours can rival the performance of sophisticated algorithms used by Onelook and \cite{hill:15}, suggesting that the information being retrieved by those algorithms is pretty basic. This calls for methods which could significantly outperform a simple approach like ours, towards encoding phrasal semantics.

Dealing with multi-word expressions isn't straightforward in our approach. We separate all words in the input phrase towards implementing our similarity measure. Detecting multi-word expressions would require recursive parsing of the phrase, something which is more suited to recurrent neural network-based approaches~\cite{hill:15}. This isn't a major concern for our task though, as the input phrase is supposed to be a simple description of the concept in mind, in which case the user is more likely to input `to die' than `kick the bucket'. Multi-word expressions are also rarely used to define words or other multi-word expressions. So, they could be treated as one node with no back-linked connections but with multiple forward-linked connections (the definition of that expression), and thus be encompassed in our approach as outputs, but not as inputs (which we do in the $80$k WordNet case).

\section{Concluding Remarks}
\label{sec:concl}

We reported the construction of a reverse dictionary based on a node-graph architecture, which derives its semantic information exclusively from dictionary definitions. The approach works at least as well as the Onelook reverse dictionary and a RNN-based approach described in~\cite{hill:15}, on a lexical size of $3$k words, but the performance deteriorates, to below Onelook's, when scaled to a lexicon with $80$k words. The performance still stays significant (as compared to the `Chance'), and greater than a forward map approach. Furthermore, this approach can be generalised to any language given an appropriate forward dictionary, lemmatizer, and a list of functional words.

Recent distributional approaches use vector representations for word meaning, derived through similarities gauged by the occurrences and co-occurrences of words in a large corpus~\cite{Erk:12}. The performance of one of these approaches, known as \emph{word2vec}~\cite{mikolov:13b,mikolov:13a}, on our test is poor, as seen in Table.~\ref{tab:performance} (under `W$2$V'). The performance suggests that phrasal semantics doesn't necessarily follow a linear additive structure.\nocite{levy:14} Indeed, researchers have been trying to find other mathematical structures and approaches which would be suitable for phrasal semantics~\cite{baroni:10,socher:11}, but with partial success and on specific types of phrases.

A class of Artificial Neural Networks (ANNs), called Recurrent Neural Networks (RNNs) are being used for tasks such as machine translation~\cite{cho:14} and generating natural image captions~\cite{karpathy:15}, among others~\cite{zhang:15}. These `deep' networks are not trained on, or to obtain, discrete syntactic categories such as NP and VP. Instead they are provided with just the inputs and expected outputs (task-dependent) while training. The learning paradigm generates features (often incomprehensible in terms of classical linguistics) on its own to effectively implement the given task\footnote{\scriptsize {``The Unreasonable Effectiveness of Recurrent Neural Networks" by Andrej Karpathy - http://karpathy.github.io/2015/05/21/rnn-effectiveness/}}, which seems to be better than using predetermined features. \cite{hill:15} use such a network to implement a reverse dictionary, and it performs at least as well as Onelook. Although the performance is noteworthy, the fact that a simple approach like ours can rival it suggests that the RNN-based approaches require further research before doing for reverse dictionaries (and phrasal semantics, in general) what Convolutional Neural Networks (CNNs) did for visual object classification~\cite{chatfield:14}. 

It seems that the focus on constituent trees and the structural combination of words cannot be compromised upon. RNNs might be the way forward, in this regard, as they could develop properties encompassing and surpassing those classical linguistic features.


\bibliography{RD16}
\bibliographystyle{emnlp2016}

\end{document}